\documentclass[pdflatex,sn-mathphys-num]{sn-jnl}
\usepackage{xcolor}
\usepackage{graphicx}
\usepackage{amssymb}
\usepackage{amsmath}
\usepackage{import}
\usepackage{pgfplots}
\usepackage{booktabs}
\usepackage{multirow}
\usepackage[table]{xcolor}
\begin{document}

\title{Beyond Dark Knowledge: Mixup-Based Distillation for Reliable Predictions}

\author*[1]{\fnm{Jos\'e} \sur{Medina}}\email{joseluis@ut.ee}

\author[2]{\fnm{Paul} \sur{Honeine}}\email{paul.honeine@univ-rouen.fr}

\author[3]{\fnm{Abdelaziz} \sur{Bensrhair}}\email{abdelaziz.bensrhair@insa-rouen.fr}

\author[1]{\fnm{Amnir} \sur{Hadachi}}\email{hadachi@ut.ee}

\affil[1]{\orgname{ITS Lab, Institute of Computer Science, University of Tartu}, \country{Estonia}}

\affil[2]{\orgname{LITIS, Universit\'e de Rouen}, \country{France}}

\affil[3]{\orgname{LITIS, INSA de Rouen}, \country{France}}

\abstract{%
Knowledge Distillation (KD) and mixup have proven effective at inducing smoothness in class boundaries; KD captures inherent class relationships in probability distributions, and mixup enforces them through convex combinations of inputs. Their interaction, however, remains poorly understood, particularly when mixup is applied only during student training. In this setting, the teacher is queried on inputs drawn from a vicinal distribution it never saw during training, a controlled mismatch whose effect on knowledge transfer has not been characterised. We show that this mismatch causes the teacher's supervisory signal to be dominated by distributional confusion rather than inter-class structure. Despite it, the student does not merely imitate the teacher: it independently acquires greater linearity in the vicinal region, a structural property that the teacher lacks, and goes beyond dark-knowledge transfer. KD with mixup consistently improves student accuracy and reduces overconfidence by an order of magnitude relative to the baseline, across CIFAR and ImageNet with varying-capacity teachers. Crucially, calibration propagates from teacher to student independently of accuracy transfer, and temperature scaling governs a measurable accuracy–calibration trade-off that becomes more pronounced under vicinal training. These results reframe mixup distillation not as a degraded version of standard KD, but as a richer transfer channel that simultaneously shapes discriminative performance, uncertainty estimation, and representational geometry.
}
\keywords{Knowledge Distillation, Mixup, Vicinal Risk Minimisation, Image Classification, Computer Vision}

\maketitle
\section{Introduction}

Knowledge Distillation (KD) is a widely adopted paradigm for transferring knowledge from a large teacher model to a smaller student model by matching soft class probability distributions \citep{hinton2015distilling,gou2021knowledge,yim2017gift}. Instead of relying solely on hard labels, the student learns from the teacher’s soft predictions, which encode inter-class relationships often referred to as \textit{dark knowledge}. This additional information enables improved generalisation and allows compact models to approach the performance of larger architectures.

In parallel, mixup has emerged as an effective data augmentation technique for improving generalisation in deep neural networks (DNN) \citep{zhang2017mixup,carratino2022on,choi2023understanding,cao2024survey}. Mixup constructs virtual training examples through convex combinations of input samples and their corresponding labels, encouraging linear behaviour between samples and leading to smoother decision boundaries. This has been shown to improve robustness and calibration of DNN predictions.

Although both KD and mixup are effective training strategies, their interaction is not straightforward. Recent work has identified \emph{smoothness} as a key factor connecting both methods and has shown that their interplay deserves explicit analysis rather than being treated as a simple additive combination \citep{choi2023understanding, guo2017calibration}. Several studies have integrated mixup into distillation frameworks to improve accuracy, robustness, or data efficiency \citep{yang2022mixskd,xu2023computation, wu2022mkd,tang2024aug,yue2025enhancing,gholami2025latent}, showing that the combination can be beneficial in practice but typically focuses on specific training schemes rather than on the effectiveness of knowledge transfer from the teacher signal.

Current literature has not systematically addressed how mixup  affects the teacher's distillation signal when vicinal inputs are introduced at inference time rather than during training. Existing work has largely prioritised accuracy \citep{yang2022mixskd, li2021smile, zhang2024hybrid}, while calibration has been treated as a secondary concern or studied independently \citep{mishra2025beyond, zhang2022and}. The information-theoretic mechanism by which mixup-induced smoothing alters the quality of knowledge transfer remains uncharacterised.

In this work, we characterise how a teacher operating on mixup-induced vicinal inputs affects the quality and structure of the distillation signal, analysing the interplay between temperature sensitivity, representation geometry, calibration, and out-of-distribution (OOD) robustness through the lens of mutual information. 
Rather than treating the teacher's behaviour under vicinal inputs as an obstacle to be mitigated, we study it as a phenomenon to be understood, identifying the conditions under which vicinal teacher inputs benefit or degrade knowledge transfer. Our empirical analysis demonstrates that students trained under such conditions acquire structural properties beyond direct imitation of the teacher, suggesting that the distillation signal carries information that transcends the teacher's in-distribution predictions. Code to reproduce all experiments is available at \href{https://github.com/JoseLMedinaC/KD-Mixup}{github.com/JoseLMedinaC/KD-Mixup}. 

\section{Background}
\label{sec:background}
\subsection{Knowledge Distillation}
KD is a model compression and knowledge transfer technique in which a smaller network is trained to mimic the predictions of a larger model. The method was formalised by \citet{hinton2015distilling}, who introduced the concept of \textit{dark knowledge}, referring to the information contained in the soft probability distribution produced by a trained classifier. Instead of training the student solely using hard labels, KD encourages the student to match the teacher’s softened output distribution. These soft targets encode inter-class relationships, which provide richer supervision.

Let $z_t$ and $z_s$ denote the logits produced by the teacher and student networks, respectively. The temperature $T$ scales the softmax distribution $p_i = \frac{\exp(z_i / T)}{\sum_j \exp(z_j / T)}$, controlling the smoothness of the probability distribution.

The distillation objective loss is a Kullback--Leibler divergence between the teacher $p_t^T$ and student $p_s^T$ T-scaled softmax outputs. 
\begin{equation}
L_{KD} =
\mathbb{E}_{x \sim P_{data}}
\left[
D_{KL}(p_t^T(x) \parallel p_s^T(x))
\right].
\label{eq:kl}
\end{equation}
\citet{hinton2015distilling} proposed that the final loss used to train a student should be a weighted combination of a distillation loss and a standard cross-entropy loss $L_{CE}$. 
\begin{equation}
    L = \alpha L_{CE} + \beta L_{KD},
    \label{eq:totalloss}
\end{equation}
\subsection{Empirical and Vicinal Risk Minimisation}
In supervised learning, the goal is to learn a function $f: \mathcal{X} \rightarrow \mathcal{Y}$ that minimises the expected risk $R(f)$ over the true distribution $P(X,Y)$. $R(f)$ is defined as $\mathbb{E}_{(X,Y)\sim P(X,Y)}[\ell(f(X),Y)]$, where $\ell$ is a loss function. 

However, the distribution $P$ is inaccessible; a learner can only access a finite training sampled dataset.  To make learning feasible, the continuous distribution is replaced by the empirical distribution $P_{\delta}(x,y) = \frac{1}{n} \sum_{i=1}^{n} \delta(x - x_i, y - y_i)$, where $\delta(\cdot)$ is the Dirac delta function centered at each training point. The resulting empirical risk is thus
\begin{equation}
    R_{\delta}(f) = \int \ell(f(x),y)\, dP_{\delta}(x,y)
                  = \frac{1}{n}\sum_{i=1}^{n} \ell(f(x_i), y_i)
    \label{eq:emprisk}
\end{equation}

Minimising this approximation constitutes the classical \textit{Empirical Risk Minimisation} (ERM) principle \citep{vapnik1999overview}. However, ERM exhibits fundamental limitations when applied to high-capacity models. In particular, since $P_{\delta}$ assigns all probability mass to isolated training points, the learned function $f$ is constrained at those fix locations, which can lead to overfitting and poor generalisation \citep{zhang2016understanding}. This behaviour reflects that the empirical distribution $P_{\delta}$ has zero support in regions between the observed samples.

This discreteness of $P_{\delta}$ is address by the \textit{Vicinal Risk Minimization} (VRM) principle \citep{chapelle2000vicinal}, which uses a vicinal distribution $P_{\nu}$,
\begin{equation}
    P_{\nu}(x,y) = \frac{1}{n}\sum_{i=1}^{n} \nu(x,y \mid x_i,y_i),
\end{equation}
that assigns probability mass to observed samples $(xi, yi)$ and their vicinities $\nu(x,y \mid x_i,y_i)$. The formulation enlarges the support of the training distribution, regularising $f$ to behave smoothly around observed data.

\subsection{ Mixup}
 Mixup is a data augmentation technique introduced to improve generalisation by encouraging DNN to behave linearly between training samples \citep{zhang2017mixup, cao2024survey}. Instead of training on individual samples, mixup constructs virtual examples by interpolating both inputs and labels. Given two training samples $(x_i, y_i)$ and $(x_j, y_j)$, mixup generates a new sample defined as:
\begin{equation}
\tilde{x} = \lambda x_i + (1-\lambda)x_j \quad , \quad 
\tilde{y} = \lambda y_i + (1-\lambda)y_j
\label{eq:xmix}
\end{equation}
where $\lambda$ sampled as $\lambda \sim \mathrm{Beta}(\alpha,\alpha)$, with $\alpha$ controlling interpolation strength.

As a VRM method, mixup deploys a continuous distribution $P_{\nu}(x,y)$ whose support spans the linear segments connecting all pairs of training examples:

\begin{equation}
P_{\nu}=
\frac{1}{n^2}\sum_{i=1}^n \sum_{j=1}^n
\mathbb E_{\lambda}
\left[
\delta\!\left(x-(\lambda x_i+(1-\lambda)x_j)\right)
\delta\!\left(y-(\lambda y_i+(1-\lambda)y_j)\right)
\right]
\end{equation}

Training under mixup forces the model to favour locally linear behaviour between training examples, effectively reducing oscillations in $f$ across classes and encouraging smoother decision boundaries \citep{zhang2017mixup, berthelot2019mixmatch, verma2019manifold, cao2024survey}. 


\section{Related Work}

\subsection{Knowledge Distillation: Paradigms and Perspectives}

Knowledge Distillation was originally introduced as a mechanism for transferring information through softened output distributions \citep{hinton2015distilling}. Subsequent work has expanded the scope of what is transferred: feature-based methods align intermediate representations between teacher and student \citep{romero2014fitnets, zhang2024hybrid}, relational approaches preserve pairwise structural similarities across samples \citep{park2019relational, yue2025enhancing}, and contrastive formulations enforce representational agreement via negative-pair objectives \citep{tian2019contrastive, zhu2025ckd}. Across these variants, a shared assumption persists: the teacher is evaluated on inputs drawn from the same distribution it was trained on, and its predictions are treated as reliable supervisory signals.

Within the dominant offline teacher-student framework  \citep{hinton2015distilling,gou2021knowledge,li2025dual,shing2025taid}, this transfer process has been formalised from an information-theoretic perspective as maximising a lower bound on the mutual information (MI) between 
teacher and student representations \citep{ahn2019variational}. A view extended to jointly estimate and maximise MI across local and global features \citep{tian2019contrastive, wang2021knowledge}, and further grounded by framing the teacher's soft outputs as an approximation of the Bayes conditional probability distribution over labels, where proximity to this distribution directly bounds student error \citep{ye2024bayes}. Together, these perspectives establish mutual information as a principled lens for evaluating the quality of knowledge transfer.

Despite the breadth of these contributions, the reliability of the teacher as a supervisory signal has received comparatively little scrutiny. Whether teacher predictions remain well-calibrated and informationally rich when the teacher operates on OOD inputs that deviate from its training distribution is a question that the existing literature has largely left open.

\subsection{Mixup as Structural Regularisation}

Mixup introduces the vicinal risk minimisation framework by training on convex combinations of inputs and labels, enforcing smoother decision boundaries and improving generalisation \citep{zhang2017mixup, cao2024survey, jin2024survey}. \citet{carratino2022on} analyse this as a mechanism that constrains model behaviour between training samples, reducing overfitting. Beyond input-level interpolation, \citet{verma2019manifold} show that applying mixup directly in hidden representations produces more compact class clusters and smoother feature geometries, suggesting that the regularisation effect operates across the full representational hierarchy of the model.

The mixup smoothing behaviour has also been linked to improved calibration \citep{jin2024survey,zhang2022and}. Calibration is particularly important in deep neural networks, which are known to produce overconfident predictions \citep{guo2017calibration}. Similar to label smoothing, which redistributes probability mass to mitigate overconfidence \citep{muller2019does}, mixup has been consistently associated with better calibrated and more robust models \citep{jin2024survey}.

However, these properties have been studied exclusively in the context of a single model trained with mixup. How the output distribution of a pre-trained model changes when it receives vicinal inputs has not been characterised.  This distinction is critical when the model in question is a frozen teacher whose predictions serve as supervisory signals for a student.

\subsection{Mixup-based Knowledge Distillation}
Motivated by its regularisation properties, several works have 
incorporated mixup into KD frameworks. \citet{beyer2022knowledge} 
show that enforcing consistency between teacher and student 
predictions on interpolated inputs improves performance through 
function matching, while \citet{yang2022mixskd} apply 
interpolation in feature space to improve representation 
alignment. \citet{xu2023computation} further leverage mixup 
to improve efficiency by focusing teacher queries on uncertain 
samples. Complementing these approaches, \citet{mishra2025beyond} 
show that mixup-based distillation can improve calibration and 
robustness under corrupted distributions, suggesting that 
vicinal samples may regularise the transferred predictive 
uncertainty.

The most systematic empirical analysis of this combination 
is provided by \citet{choi2023understanding}, who demonstrate 
that a mixup-trained teacher produces degraded distillation 
signals due to excessive smoothness in features and logits, 
and propose practical remedies including rescaled logits and 
partial mixup. Their findings establish that the teacher's 
training regime directly affects the quality of knowledge 
transfer, and that temperature interacts non-trivially with 
mixup-induced smoothness.

Despite these contributions, the existing literature treats 
the teacher's behaviour under vicinal inputs as a practical 
obstacle to be mitigated rather than a phenomenon to be 
understood. None of these works characterise how mixup-induced 
changes in the teacher's output distribution affect the mutual 
information transferred to the student, nor how these changes 
propagate into student calibration and generalisation.

\subsection{Distillation under Vicinal Distributions}

A separate line of research has examined KD beyond 
in-distribution supervision. \citet{fang2021mosaicking} showed that OOD data can augment distillation when samples are first mapped toward the teacher's training distribution, recovering reliable supervisory signals without access to the original data. \citet{tang2024aug} take a different approach, anchoring interpolations between OOD and in-domain samples to improve robustness under domain shift. 
More recently, \citet{zhang2025shiftkd} 
systematically benchmarked over thirty KD methods under 
diversity and correlation shift, revealing substantial 
degradation across the board, and \citet{stanton2021does} 
demonstrated that closely mimicking the teacher does not 
guarantee student generalisation even under standard 
in-distribution conditions.

Critically, these works study distribution shift as a 
property of the data environment in which the student 
learns, not as a property of the inputs the teacher 
receives at inference time. The question of whether teacher predictions remain reliable supervisory signals when the teacher itself operates on vicinal inputs drawn from outside its training distribution has not been directly addressed. \citet{choi2023understanding} 
provide the closest prior work, identifying empirically 
that vicinal teacher inputs degrade the distillation 
signal, yet their analysis stops at the observational 
level. The information-theoretic mechanism by which 
mixup-induced inputs alter mutual information transfer, 
teacher calibration, and student generalisation under 
distribution shift remains uncharacterised.

This work addresses that gap by explicitly analysing how 
a teacher operating on vicinal inputs affects the quality 
of knowledge transfer, measured through mutual information, 
calibration, and robustness, and by identifying the 
conditions under which such out-of-distribution teacher 
behaviour benefits or degrades student learning.

\section{Problem Formulation}
\label{sec:problem}
\begin{figure} [h!]
    \centering
    \def\svgwidth{0.7\columnwidth}
    \begingroup
        \fontsize{5}{5}\selectfont
        \import{images/}{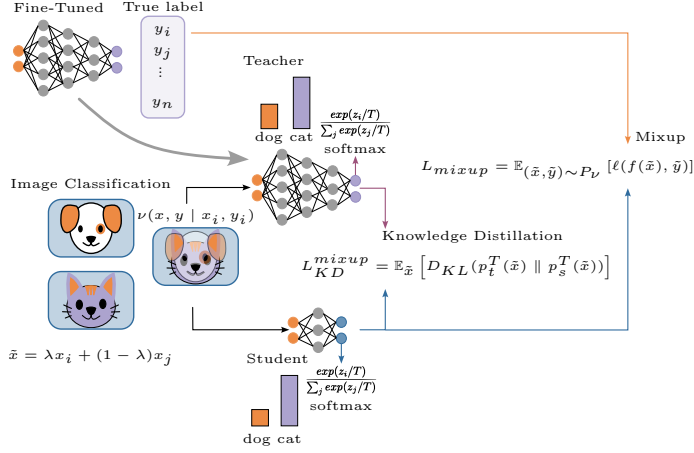}
    \endgroup
    \caption{Mixup-based knowledge distillation pipeline. Interpolated samples $\tilde{x}=\lambda x_i + (1-\lambda)x_j$ are generated and evaluated by both teacher and student. The student is trained using a combination of mixup supervision and distillation loss $\mathrm{KL}(p_t^T(\tilde{x}) \| p_s^T(\tilde{x}))$, while the teacher remains fixed and is queried on vicinal inputs, introducing a distribution mismatch between $P_{\text{data}}$ and $P_{\nu}$.}
    \label{fig:mixup}
\end{figure}

In this section, we formalise the interaction between KD and mixup applied during the student training stage, following the scheme illustrated in Figure~\ref{fig:mixup}, where the teacher is queried on $\tilde x$ for distillation. All subsequent analysis is based on this setting. 

\begin{figure}
    \centering
    \def\svgwidth{0.75\columnwidth}
    \begingroup
        \fontsize{4}{4}\selectfont
\begingroup%
  \makeatletter%
  \providecommand\color[2][]{%
    \errmessage{(Inkscape) Color is used for the text in Inkscape, but the package 'color.sty' is not loaded}%
    \renewcommand\color[2][]{}%
  }%
  \providecommand\transparent[1]{%
    \errmessage{(Inkscape) Transparency is used (non-zero) for the text in Inkscape, but the package 'transparent.sty' is not loaded}%
    \renewcommand\transparent[1]{}%
  }%
  \providecommand\rotatebox[2]{#2}%
  \newcommand*\fsize{\dimexpr\f@size pt\relax}%
  \newcommand*\lineheight[1]{\fontsize{\fsize}{#1\fsize}\selectfont}%
  \ifx\svgwidth\undefined%
    \setlength{\unitlength}{2152.68016425bp}%
    \ifx\svgscale\undefined%
      \relax%
    \else%
      \setlength{\unitlength}{\unitlength * \real{\svgscale}}%
    \fi%
  \else%
    \setlength{\unitlength}{\svgwidth}%
  \fi%
  \global\let\svgwidth\undefined%
  \global\let\svgscale\undefined%
  \makeatother%
  \begin{picture}(1,0.34226092)%
    \lineheight{1}%
    \setlength\tabcolsep{0pt}%
    \put(0,0){\includegraphics[width=\unitlength,page=1]{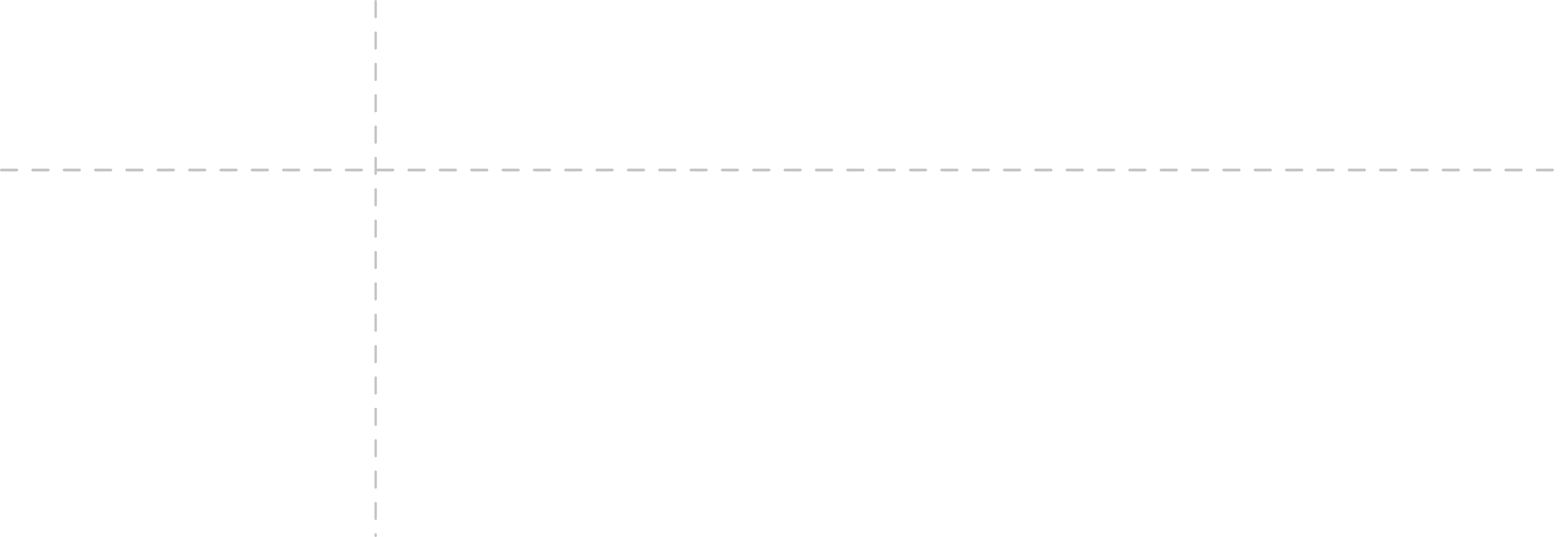}}%
    \put(0.11598451,0.31342917){\color[rgb]{0,0,0}\makebox(0,0)[t]{\lineheight{1.10000002}\smash{\begin{tabular}[t]{c}Empirical \\Risk\\Minimisation\end{tabular}}}}%
    \put(0.60609294,0.31342917){\color[rgb]{0,0,0}\makebox(0,0)[t]{\lineheight{1.10000002}\smash{\begin{tabular}[t]{c}Visinal \\Risk\\Minimisation\end{tabular}}}}%
    \put(0.34510215,0.0417475){\color[rgb]{0,0,0}\makebox(0,0)[t]{\lineheight{1.10000002}\smash{\begin{tabular}[t]{c}Mixup\end{tabular}}}}%
    \put(0.60898292,0.04157329){\color[rgb]{0,0,0}\makebox(0,0)[t]{\lineheight{1.10000002}\smash{\begin{tabular}[t]{c}Knowledge \\Distillation\end{tabular}}}}%
    \put(0,0){\includegraphics[width=\unitlength,page=2]{empircal_visinal.pdf}}%
    \put(0.86717348,0.04157329){\color[rgb]{0,0,0}\makebox(0,0)[t]{\lineheight{1.10000002}\smash{\begin{tabular}[t]{c}Mixup-based\\Distillation\end{tabular}}}}%
  \end{picture}%
\endgroup%

    \endgroup
    \caption{Effect of ERM and VRM on decision boundaries. Under ERM, the model produces sharp, confident separations. Under VRM, three cases are shown: mixup alone softens the boundary into a continuous transition region; Knowledge Distillation introduces smoothness through temperature-scaled supervision; and mixup-based distillation combines both effects, producing a boundary that reflects the interaction between vicinal training and soft-label transfer. Balancing these two sources of smoothing is the central challenge this work addresses.}
    \label{fig:emp_visi}
\end{figure}
Before formalising the distillation setting, we briefly motivate the need for a controlled balance between the two smoothing mechanisms. Figure \ref{fig:emp_visi} illustrates how ERM and VRM affect decision boundaries. Under ERM, the model learns sharp, highly confident boundaries. Mixup, as a VRM method, replaces these with smooth transition regions rather than hard separations, improving generalisation but reducing discriminative sharpness. Knowledge Distillation introduces a different form of softness through temperature-scaled logits, encouraging the student to capture inter-class structure from the teacher. When both mechanisms are combined, as in mixup-based distillation, the resulting boundary reflects a balance between these two sources of smoothing. Understanding and controlling this balance is the central concern of this work. A deeper analysis of how each regime affects feature clustering geometry, including t-SNE visualisations across architectures and training regimes, is provided in the appendix \ref{app:capacity_smootnes}.
\subsection{Distillation Setting}

Consider a classification problem with input space $\mathcal{X}$ and label space $\mathcal{Y} = \{1,\dots,K\}$. Let $D = \{(x_i,y_i)\}_{i=1}^{N}$, denote a dataset sampled from a data distribution $(x,y)$. In mixup-based KD, we assume i) access to a pretrained teacher trained using $D$, ii) a mix training dataset sampled as $(\tilde{x},\tilde{y}) \sim P_{\nu}$, and iii) a student training based on soft targets. Let $z_t(\tilde{x})$ and $z_s(\tilde{x})$ denote the logits produced by the teacher and student, respectively. The temperature-scaled softmax probabilities are defined for $(\tilde{x},\tilde{y})$ as:
\begin{equation}
p_t^T(\tilde{x})_i =
\frac{\exp(z_{t,i}(\tilde{x})/T)}
{\sum_{j=1}^{K}\exp(z_{t,j}(\tilde{x})/T)}, \quad
p_s^T(\tilde{x})_i =
\frac{\exp(z_{s,i}(\tilde{x})/T)}
{\sum_{j=1}^{K}\exp(z_{s,j}(\tilde{x})/T)}.
\end{equation}

\subsection{Mixup and Vicinal Training Distribution}
Sampling $(\tilde{x},\tilde{y}) \sim P_{\nu}$ as equation \ref{eq:xmix} corresponds to constructing the mixup by selecting pairs of training examples and a mixing coefficient $\lambda$. Under this formulation, the learning objective becomes $L_{mixup} =
\mathbb{E}_{(\tilde{x},\tilde{y})\sim P_{\nu}}
\left[
\ell(f(\tilde{x}),\tilde{y})
\right]$, expanded as:
\begin{equation}
L_{mixup}
=
\frac{1}{n^2}\sum_{i=1}^{n}\sum_{j=1}^{n}
\mathbb{E}_{\lambda}
\left[
\ell\!\left(f\!\left(\lambda x_i+(1-\lambda)x_j\right),\;
\lambda y_i+(1-\lambda)y_j\right)
\right].
\end{equation}


Therefore under the same $P_{\nu}$, the distillation objective in equation~\ref{eq:kl} changes and becomes:
\begin{equation}
L_{KD}^{mixup} =
\mathbb{E}_{\tilde{x} \sim P_{\nu}}
\left[
D_{KL}(p_t^T(\tilde{x}) \parallel p_s^T(\tilde{x}))
\right]
\end{equation}

Since the teacher was not trained on the vicinal distribution $P_{\nu}$, its predictions on interpolated inputs may exhibit different uncertainty properties \citep{guo2017calibration, mishra2025beyond}. However, these two training losses encourage the student model to behave linearly between training examples and typically lead to smoother decision boundaries.  

\subsubsection{Information-Theoretic Perspective}
\label{subsubsec:mi}
 
Knowledge distillation can be understood as maximising a lower bound
on the mutual information between teacher and student
predictions~\citep{ahn2019variational, tian2019contrastive}:
\begin{equation}
  I(p_t;\, p_s)
  \;\geq\;
  H\!\left[p_t(x)\right]
  \;+\;
  \mathbb{E}_{x}\!\left[\log p_s^T(x)\right],
  \label{eq:mi_bound}
\end{equation}
where $H[p_t(x)]$ is the entropy of the teacher's predictions and
$p_s^T(x)$ denotes the temperature-scaled softmax output of the
student.
Since the teacher is fixed and evaluated on real inputs
$x \sim \mathcal{P}_\text{data}$, the term $H[p_t(x)]$ is constant
with respect to the student parameters. Minimising $\mathcal{L}_{KD}$ is therefore equivalent to maximising the lower bound (equation \eqref{eq:mi_bound}). The lower bound provides a minimum guarantee of the amount of information transferred when a student learns to reproduce a teacher signal as faithfully as possible.
 
In our setting, however, the teacher is queried on interpolated inputs
$\tilde{x}$, which breaks the assumption that $H[p_t(x)]$ is constant. Thus, the teacher entropy becomes:
\begin{equation}
  H\!\left[p_t(\tilde{x})\right]
  \;=\;
  H\!\left[p_t(x)\right] + \mathcal{L}_\text{NL}(\lambda),
  \label{eq:entropy_decomp}
\end{equation}
where $\mathcal{L}_\text{NL}(\lambda)$ captures the additional entropy
introduced by distributional mismatch rather than by the natural
ambiguity of the mixed label.
Consequently, the lower bound grows with $\mathcal{L}_\text{NL}$, but this growth does not reflect an
increase in useful information available to the student; it reflects the teacher's confusion when evaluated outside its training distribution.

\subsection{Teacher Reliability}
 
The effectiveness of knowledge distillation depends on the reliability of the teacher's predictions when evaluated on the inputs presented during student training with interpolated inputs $\tilde{x} \sim \mathcal{P}_\nu$ \citep{guo2017calibration}. We characterise teacher reliability with respect to vicinal inputs using two complementary quantities.
 
The first is the \emph{non-linearity residual} $\mathcal{L}_\text{NL}(\lambda)$, which measures the discrepancy between the teacher’s prediction on an interpolated input $\tilde{x} = \lambda x_i + (1{-}\lambda) x_j$ and the prediction of a perfectly linear model, namely $\lambda\, p_t(x_i) + (1{-}\lambda)\, p_t(x_j)$. It is defined as

\begin{equation}
  \mathcal{L}_\text{NL}(\lambda)
  \;=\;
  \mathbb{E}_{i,j}\!\left[
    D_\text{KL}\!\left(
      \lambda\, p_t(x_i) + (1{-}\lambda)\, p_t(x_j)
      \;\Big\|\;
      p_t(\tilde{x})
    \right)
  \right],
  \label{eq:lnl}
\end{equation}
In other words, we are measuring how the teacher deviates from linear behaviour between training examples when evaluated
on different $\lambda$ interpolated inputs. 
A teacher with $\mathcal{L}_\text{NL} = 0$ responds perfectly linearly
to convex combinations of its training samples; larger values indicate
that the teacher's predictions on $\tilde{x}$ are inconsistent with its
predictions on the individual samples $x_i$ and $x_j$.
 
The second is the \emph{dominance ratio} $\mathcal{D}(\lambda)$, defined
as
\begin{equation}
  \mathcal{D}(\lambda)
  \;=\;
  \frac{\mathcal{L}_\text{NL}(\lambda)}{H\!\left[p_t(\tilde{x})\right]},
  \label{eq:dominance}
\end{equation}
which quantifies what fraction of the total entropy \citep{shannon1948mathematical}  of the teacher's prediction on $\tilde{x}$ is attributable
to the distributional mismatch rather than to the natural ambiguity of the mixed label.
When $\mathcal{D}(\lambda) > 1$, the mismatch term exceeds the total prediction entropy, indicating that the supervisory signal is dominated by the teacher's confusion rather than by informative inter-class structure. Together, $\mathcal{L}_\text{NL}$ and $\mathcal{D}$ characterise complementary aspects of teacher reliability on vicinal inputs. 

A third quantity captures what the student retains relative to the teacher. We define the \emph{linearity information}
of a model $f$ on vicinal inputs as %
\begin{equation}
  \mathcal{I}_\text{lin}(f, \lambda)
  \;=\;
  H[p_\text{linear}(\tilde{x})]
  \;-\;
  D_\text{KL}(p_\text{linear} \| p_f(\tilde{x})),
  \label{eq:ilin}
\end{equation}
where $p_\text{linear}(\tilde{x}) = \lambda p_f(x_i) + (1{-}\lambda) p_f(x_j)$ is the prediction a perfectly linear model
would produce. $\mathcal{I}_\text{lin}$ measures how much information of the ideal
linear prediction is retained by $f$ on the interpolated input. A perfectly linear model achieves $D_\text{KL} = 0$ and retains all of $H[p_\text{linear}]$. Since neither teacher nor student are perfectly linear on vicinal inputs, we define the \emph{linearity gain} between teacher and student as:
\begin{equation}
  \Delta_\text{lin}(\lambda)
  \;=\;
  \mathcal{L}_\text{NL}^{(t)}(\lambda)
  \;-\;
  \mathcal{L}_\text{NL}^{(s)}(\lambda),
  \label{eq:lingain_info}
\end{equation}
which is verified empirically in Section~\ref{sec:beyond}. A positive $\Delta_\text{lin}$ means the student retains more information of the ideal linear prediction than its teacher. 

\subsection{Hypothesis}
\label{sec:hypothesis}
Based on the previous formulation, we investigate the following hypothesis:
 
\begin{quote}
\textit{%
Applying Mixup during distillation introduces a vicinal distribution shift in the inputs used to query the teacher, causing the teacher's supervisory signal to be dominated by distributional mismatch rather than by informative inter-class structure. Despite this, the student does not merely imitate the teacher: it acquires a structural property, greater linearity in the vicinal region, that the teacher lacks. This property, induced by the vicinal training objective, generalises beyond the interpolation manifold and manifests as improved robustness under smooth distribution shifts.
}
\end{quote}

Our empirical study aims to analyse how mixup regularises the student toward greater linearity
independently of the teacher, how this affects the reliability and calibration of the teacher's predictions, and how these effects propagate through the distillation process to influence student performance, even when distributions are corrupted. 

\section{Experimental Setup}

In this section, we describe the experimental protocol for evaluating knowledge distillation under mixup. 

\subsection{Datasets}

We evaluate the proposed setting on three image classification benchmarks: CIFAR-10, CIFAR-100, and ImageNet. CIFAR is used as a standard low-resolution benchmark for analysing distillation and calibration, while ImageNet is included to evaluate whether the same behaviour persists in a large-scale setting.
\subsection{Teacher and Student Architectures}

We consider four teacher architectures with different capacities: ConvNeXt-Large, Vision Transformer Base with patch size 16 (ViT-B/16), ResNet152V2, and ConvNeXt-Tiny.
For all experiments, the student network is a customised, lightweight version of \textbf{MobileNet}. By replacing the standard 1,280-unit classification head with a 256-unit bottleneck layer, we further reduce the total parameters to approximately 2.85M (ImageNet), 2.61M (CIFAR-100), and 2.59M (CIFAR10). This streamlined architecture is specifically designed to evaluate the effectiveness of knowledge transfer in resource-constrained scenarios.
\subsubsection{Teacher Initialisation and Fine-Tuning}

Teacher networks are initialised from pretrained weights obtained on large-scale data (ImageNet 1K), and then fine-tuned on the target dataset. In particular, the teachers are not trained from scratch. This choice is important because it places the teacher in a realistic transfer learning setting, where pretrained representations may already provide strong semantic structure before fine-tuning.

\subsection{Distillation Setting}
We adopt the standard logit-based distillation framework described in Section~\ref{sec:background} and focus primarily on the distillation signal during training. In particular, we set $\alpha = 0$ in equation~\ref{eq:totalloss} for CIFAR experiments, removing the contribution of the cross-entropy term with ground-truth labels. This choice isolates knowledge transfer from teacher to student, allowing us to analyse how information propagates through temperature-scaled soft targets under mixup inputs, without interference from hard-label supervision. For ImageNet, however, we use a small contribution $\alpha = 0.2$ to provide limited supervision and mitigate the increased complexity and mismatch, while still preserving the dominant effect of mixup-KD. As a result, all reported experiments reflect primarily distillation-driven learning dynamics.

To study the interaction between temperature scaling and mixup interpolation, we perform a validation sweep over temperatures $T \in [2,10]$ and $T=\sigma$. Based on this analysis, we fix $T=2$ for all subsequent experiments, as it provides the best trade-off between accuracy and training stability.

\subsection{Mixup During Student Training}
Mixup is applied \emph{only during student training}. The teacher remains fixed and is queried on interpolated inputs generated online during optimisation.

We sample $\lambda \sim \mathrm{Uniform}(0,1)$ to uniformly cover the full range of interpolation strengths. This choice is motivated by our goal of evaluating teacher reliability across the entire vicinal space: by assigning equal probability to all interpolation levels, we obtain a comprehensive view of how input ambiguity affects the linearity of the teacher's predictions.

\subsection{Evaluation Metrics}

We evaluate both teacher and student models using three metrics:

\paragraph{Accuracy.}
Classification performance is measured by top-1 accuracy: $\mathrm{Acc} =
\frac{1}{N}\sum_{i=1}^{N}\mathbf{1}(\hat{y}_i = y_i)$.
\paragraph{Average Confidence.}
To quantify the confidence level of the model, we compute the mean of the maximum predicted probability over the evaluation set, $\mathrm{Conf} = \frac{1}{N}\sum_{i=1}^{N}\max_k p(k \mid x_i)$. This metric summarises how confident the model is on average, independent of whether those predictions are correct.
\paragraph{Expected Calibration Error (ECE).}
To measure calibration, we use the Expected Calibration Error (ECE):
\begin{equation}
\mathrm{ECE} =
\sum_{m=1}^{M}\frac{|B_m|}{N}
\left|
\mathrm{acc}(B_m) - \mathrm{conf}(B_m)
\right|,
\label{eq:ece}
\end{equation}

where $B_m$ is the set of predictions whose confidence falls in bin $m$, $\mathrm{acc}(B_m)$ is the empirical accuracy in that bin, and $\mathrm{conf}(B_m)$ is the average confidence. ECE measures the discrepancy between confidence and correctness. Lower ECE indicates better calibration.

\subsection{Reliability Perspective}
 
Building on the formulation introduced in Section~\ref{sec:problem}, where teacher reliability on vicinal
inputs is characterised through the non-linearity residual $\mathcal{L}_\text{NL}(\lambda)$ and the dominance ratio
$\mathcal{D}(\lambda)$, we design a set of experiments to empirically validate these measures and analyse their consequences for the distilled student.
 
\paragraph{Non-linearity and signal quality.}
For each teacher architecture, we compute $\mathcal{L}_\text{NL}(\lambda)$
and $\mathcal{D}(\lambda)$ over the training set. Results are reported as curves over $\lambda \in (0,1)$, allowing direct inspection of how the mismatch and its dominance over the supervisory signal vary with interpolation strength.
 
\paragraph{Student linearity and alignment.}
To assess what the student learns beyond imitating the teacher, we
compute $\mathcal{L}_\text{NL}(\lambda)$ for the student under the same
protocol and define the linearity gain as equation \ref{eq:lingain_info}. 

We further measure teacher--student alignment via $D_\text{KL}(p_t(\tilde{x}) \| p_s(\tilde{x}))$ on both real and interpolated inputs to quantify how much the student diverges from the
teacher in the vicinal region.
 
\paragraph{Robustness under distribution shift.}
To assess whether the linearity induced by mixup generalises beyond the
training distribution, we evaluate all teacher--student pairs on CIFAR-100-C~\citep{hendrycks2019robustness}, which provides six corruption types: gaussian noise, motion blur, fog, jpeg compression, contrast, and pixelate (each at five severity levels).
Performance is reported as accuracy gain over the baseline student $\Delta_\text{acc} = \text{acc}_\text{student} -\text{acc}_\text{baseline}$, isolating the contribution of distillation from the regularisation effect of mixup alone.

\section{Results and Discussion}
In this section, we jointly present and analyse the behaviour of teacher and student models across CIFAR and ImageNet and different teacher architectures. 

\begin{table}[t]
\caption{Accuracy--calibration trade-off under mixup-based distillation across temperatures and teacher architectures on CIFAR-100 for a student MobileNetV2. ECE decreases monotonically with temperature while accuracy degrades, revealing a consistent trade-off across all architectures. KD Classic$^*$ uses standard KD without mixup at $T{=}2$ as a reference baseline.}
\label{tab:teacher_temp_acc_1}
\centering
\setlength{\tabcolsep}{6pt}
\begin{tabular}{llrrr}
\toprule
Teacher & $T$ & Acc (\%) & Conf (\%) & ECE \\
\midrule
 & Classic Training&79.52 &--&--\\
 & Mixup Classic& 81.05&91.2&0.048\\
\midrule
\rowcolor{gray!10}\multirow{7}{*}{ConvNeXt-L} & $T{=}2$ & 86.30 & 87.52 & 0.0203 \\
 & $T{=}4$ & 85.96 & 87.25 & 0.0215 \\
 & $T{=}6$ & 85.94 & 86.01 & 0.0196 \\
 & $T{=}8$ & 85.28 & 85.27 & 0.0166 \\
 & $T{=}10$ & 85.42 & 84.36 & 0.0175 \\
 & $T_{\sigma}$ & 85.81 & 82.69 & 0.0436 \\
 & KD Classic$^*$ & 82.83 & 86.36 & 0.040 \\
\midrule
\rowcolor{gray!10}\multirow{7}{*}{ViT-B/16} & $T{=}2$ & 85.80 & 88.76 & 0.0315 \\
 & $T{=}4$ & 85.67 & 88.48 & 0.0344 \\
 & $T{=}6$ & 85.62 & 87.47 & 0.0223 \\
 & $T{=}8$ & 85.73 & 86.62 & 0.0221 \\
 & $T{=}10$ & 85.47 & 86.06 & 0.0159 \\
 & $T_{\sigma}$ & 85.69 & 81.25 & 0.0535 \\
 & KD Classic$^*$ & 82.75 & 87.44 & 0.0479 \\
\midrule
\rowcolor{gray!10}\multirow{7}{*}{ResNet152V2} & $T{=}2$ & 84.10 &88.87 & 0.0496 \\
 & $T{=}4$ & 83.83 & 88.92 & 0.0542 \\
 & $T{=}6$ & 83.68 & 88.26 & 0.0466 \\
 & $T{=}8$ & 83.52 & 87.42 & 0.0420 \\
 & $T{=}10$ & 83.51 & 86.90 & 0.0370 \\
 & $T_{\sigma}$ & 84.04 & 84.33 & 0.0356 \\
 & KD Classic$^*$ & 81.97 & 87.98 & 0.061 \\
\midrule
\rowcolor{gray!10}\multirow{7}{*}{ConvNeXt-T} & $T{=}2$ & 81.10 & 77.48 & 0.0364 \\
 & $T{=}4$ & 81.01 & 76.47 & 0.0454 \\
 & $T{=}6$ & 80.50 & 74.65 & 0.0584 \\
 & $T{=}8$ & 80.31 & 73.94 & 0.0637 \\
 & $T{=}10$ & 79.77 & 73.05 & 0.0672 \\
 & $T_{\sigma}$ & 81.02 & 80.14 & 0.019 \\
 & KD Classic$^*$ & 79.63 & 76.56 & 0.031 \\
\bottomrule
\end{tabular}
\end{table}
\subsection{Effect of Temperature under Mixup Distillation}

Table~\ref{tab:teacher_temp_acc_1} shows the impact of the distillation temperature under mixup-based student training. Three consistent patterns emerge:

\textbf{(1) Mixup improves the classic training student.}  
The baseline student improves from $79.52\%$ to $81.05\%$, confirming the regularisation effect of vicinal training \citep{zhang2017mixup} even without teacher supervision.

\textbf{(2) Distillation provides additional gains.}  
All teacher-student configurations outperform the mixup classic and KD classic baselines, with gains ranging from $\sim$0.5\% (ConvNeXt-Tiny) to $\sim$5.2\% (ConvNeXt-Large), indicating that mixup and KD act as complementary mechanisms.

\textbf{(3) Moderate temperature is critical.}  
Across all teachers, $T=2$ yields the best performance. Increasing $T$ consistently degrades accuracy, particularly for weaker teachers, suggesting that excessive softening removes informative structure from the teacher logits. The rescaled logit normalisation $T=\sigma$ proposed by \cite{choi2023understanding} achieved competitive accuracy but was still outperformed by $T=2$.

\textbf{(4) Mixup improves calibration independently of temperature.} Across all temperatures, KD+mixup achieves lower ECE than standard KD without mixup. This confirms that mixup is the primary driver of calibration improvement, reducing overconfidence structurally through vicinal training rather than through soft-label smoothing alone.

\textbf{(5) Temperature controls an explicit accuracy--calibration trade-off.} Within KD+mixup, ECE decreases monotonically with temperature while accuracy degrades consistently across all teacher architectures. This reveals that temperature is not a neutral hyperparameter under vicinal distillation: it governs a measurable trade-off between discriminative performance and uncertainty estimation. $T=2$ optimises accuracy at the cost of calibration, while higher temperatures improve ECE at the cost of accuracy. $T_{\sigma}$ represents the pathological case, producing the worst ECE despite the lowest average confidence, confirming that logit normalisation does not equivalent calibration.\\

A complementary analysis of temperature and its effects during training is shown and explained in Appendix \ref{sec:temp_effects}

\begin{table} [h!]
\caption{Performance comparison on CIFAR-10. We report top-1 accuracy, average confidence, and ECE for teacher and student models. Distillation with Mixup improves accuracy over the baseline student while significantly reducing overconfidence (lower ECE). Student calibration closely follows teacher reliability.}
\centering
\begin{tabular}{lccc}
\hline
Model & Accuracy & Confidence & ECE \\
\hline
ConvNeXt-Large (Teacher) & 0.9856 & 0.9789 & 0.0070 \\
ViT-B/16 (Teacher)  & 0.9850 & 0.9927 &  0.0085 \\
ResNet152V2 (Teacher) & 0.9698 & 0.9838 & 0.0152 \\
ConvNeXt-Tiny (Teacher) & 0.9672 & 0.9744 & 0.0104 \\
\hline
MobileNetV2 (KD + mixup, ConvNeXt-L) & 0.9760 & 0.9726 & 0.0051 \\
MobileNetV2 (KD + mixup, ViT-B/16) & 0.9780 & 0.9859 & 0.0094 \\
MobileNetV2 (KD + mixup, ResNet152V2) & 0.9733 & 0.9848 & 0.0125 \\
MobileNetV2 (KD + mixup, ConvNeXt-T) & 0.9697 & 0.9763 & 0.0072 \\
\hline
MobileNetV2 Baseline & 0.9635 & 0.9863 & 0.1073 \\
\hline
\end{tabular}
\label{tab:cifar10result_1}
\end{table}
\subsection{Results on CIFAR-10}
Table~\ref{tab:cifar10result_1} shows that all teachers are highly accurate and well-calibrated (ECE $< 0.02$). Distillation improves the baseline from $96.35\%$ to up to $97.60\%$, suggesting that the student approaches an empirical capacity ceiling. Despite being a lower-capacity architecture, the MobileNet student can match or even slightly surpass the performance of some teachers trained under standard fine-tuning regimes. This suggests that mixup-based KD enables a more efficient utilisation of the student’s representational capacity, allowing it to approach its architectural limit. 

The baseline is strongly overconfident (ECE $=0.1073$), while distillation reduces ECE by an order of magnitude across all teachers. Better-calibrated teachers (ConvNeXt-Large) yield the lowest student ECE. Less-calibrated teachers (ResNet152V2) transfer residual calibration error compared to other students, but reduce it from the teacher while also maintaining good accuracy transfer. 

In this low-complexity regime, KD with mixup enables the student to efficiently exploit its capacity, achieving near-teacher performance while significantly improving calibration.

\begin{table}
\centering
\small
\caption{Performance comparison on CIFAR-100. We report top-1 accuracy, average confidence, and ECE for teacher and student models. Accuracy gains scale with teacher strength, while calibration of the student reflects the teacher’s reliability, with better-calibrated teachers yielding lower ECE.}
\begin{tabular}{lccc}
\hline
Model & Accuracy & Confidence & ECE \\
\hline
ConvNeXt-Large (Teacher) & 0.9217 & 0.9189 & 0.0049 \\
ViT-B/16 (Teacher) & 0.9151 & 0.9272 & 0.0174 \\
ResNet152V2 (Teacher) & 0.8257 & 0.8873 & 0.0619 \\
ConvNeXt-Tiny (Teacher) & 0.8196 & 0.7908 & 0.0288 \\

\hline
MobileNetV2 (KD + MixUp, ConvNeXt-L) & 0.8630 & 0.8752 & 0.0203 \\
MobileNetV2 (KD + MixUp, ViT-B/16) & 0.8580 & 0.8876 & 0.0315 \\
MobileNetV2 (KD + MixUp, ResNet152V2) & 0.8410 & 0.8887 & 0.0496 \\
MobileNetV2 (KD + MixUp, ConvNeXt-T) & 0.8110 & 0.7748 & 0.0364 \\
\hline
MobileNetV2 Baseline & 0.8105 & 0.9527 & 0.1421 \\
\hline
\end{tabular}
\label{tab:cifar100result_1}
\end{table}
\subsection{Results on CIFAR-100}

Table~\ref{tab:cifar100result_1} highlights a clear dependency between teacher quality and student performance. Accuracy gains follow teacher strength: $+5.2\%$ (ConvNeXt-Large), $+3.75\%$ (ViT-B/16), $+3.0\%$ (ResNet152V2), and marginal improvement for ConvNeXt-Tiny. The student approaches a ceiling around $86.3\%$, remaining below the strongest teacher, indicating MobileNet's architectural constraints. For instance, the ConvNeXt-Large student ($acc=86.28\%$) is forced to reach this limit, at the cost of higher ECE; and the same for ViT-B/16.

When the teacher operates closer to the student's capacity, such as ResNet152V2, the student can better match the teacher and achieve a relative improvement for both accuracy ($+1.52\%$) and ECE ($-0.012$). This suggests that when the teacher lies within the student’s representational range, knowledge transfer becomes more effective.

The baseline student is highly overconfident (ECE $=0.1421$). Mixup-distillation significantly reduces this gap, with the best calibration achieved when distilling from ConvNeXt-Large.
Students tend to adopt their teachers' confidence profiles. In particular, underconfident teachers (ConvNeXt-Tiny) produce underconfident students, while overconfident teachers (ResNet152V2) produce improvement but still biased students.

\begin{table}[h]
\centering
\small
\caption{Performance comparison on ImageNet. We report top-1 accuracy, average confidence, and ECE for teacher and student models. Distillation yields modest accuracy gains under higher data complexity, while calibration improvements depend on teacher quality and are not consistently preserved.}
\begin{tabular}{lccc}
\hline
Model & Accuracy & Confidence & ECE \\
\hline
ConvNeXt-Large (Teacher) & 0.8096 & 0.8390 & 0.050 \\
ViT-B/16 (Teacher) & 0.8005 & 0.8275 & 0.063 \\
ResNet152V2 (Teacher) & 0.7723 & 0.8142 & 0.080 \\
ConvNeXt-Tiny (Teacher) & 0.6590 & 0.7028 & 0.152 \\

\hline
MobileNetV2 (KD + MixUp, ConvNeXt-L) & 0.7160 & 0.7331 & 0.046 \\
MobileNetV2 (KD + MixUp, ViT-B/16) & 0.7158 & 0.7329 & 0.042 \\
MobileNetV2 (KD + MixUp, ResNet152V2) & 0.7152 & 0.7362 & 0.038 \\
MobileNetV2 (KD + MixUp, ConvNeXt-T) & 0.6947 & 0.7570 & 0.194 \\
\hline
MobileNetV2 Baseline & 0.7006 & 0.8600 & 0.127 \\
\hline
\end{tabular}
\label{tab:imagenetresult_1}
\end{table}
\subsection{Results on ImageNet}

Table~\ref{tab:imagenetresult_1} extends the analysis to a higher-complexity regime. Distillation yields modest gains ($\approx $1.5\%) for strong teachers, while weak teachers show limited improvement. The gap between teacher and student remains large, indicating that student capacity becomes a dominant bottleneck.

The baseline is again overconfident (ECE $=0.127$). Distillation reduces this gap for strong teachers (ECE $\approx 0.04$). ConvNeXt-Large and ResNet152V2, which exhibit relatively good calibration, produce well-calibrated students. In contrast,  although the student distilled from ConvNeXt-Tiny improves in accuracy relative to its teacher, this gain comes at the cost of significantly worse calibration  (ECE = 0.194). 

In this regime, calibration emerges as a secondary property that is not explicitly optimised by the distillation objective. As a result, it is neither consistently preserved nor systematically improved, and may instead be propagated or even amplified under complex data settings. Notably, we observe a decoupling between accuracy and calibration; while the distillation process continues to transfer discriminative structure and improve predictive performance, it fails to maintain reliable uncertainty estimates.
\section{Beyond Dark Knowledge: What Mixup Transfers}
\label{sec:beyond}
\subsection{Calibration Behaviour under Mixup Distillation}
\begin{figure} [h!]
    \centering
    \def\svgwidth{0.78\columnwidth}
    \begingroup
        \fontsize{5}{5}\selectfont
        \import{images/}{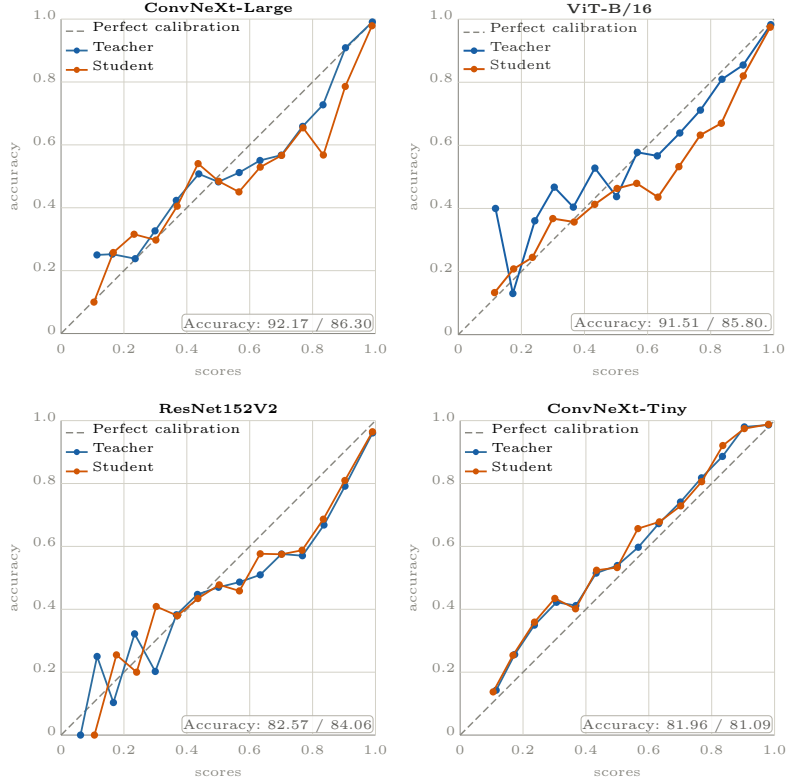}
    \endgroup
      \caption{Confidence–accuracy curves on CIFAR-100. Each panel compares a teacher with its distilled student. The diagonal represents perfect calibration. Well-calibrated teachers transfer both accuracy and calibration, overconfident teachers are partially corrected by mixup, and weak teachers limit accuracy gains while preserving calibration.}
    \label{fig:conf_acc_all_models}
\end{figure}

Figure~\ref{fig:conf_acc_all_models} compares the confidence--accuracy curves of teacher and student models. We observe three distinct regimes depending on the teacher quality:

\textbf{(1) Strong and well-calibrated teacher.}  
The student closely matches both accuracy and calibration across all confidence levels for ConvNeXt-Large, and improves underconfidence low values for ViT-B/16. This indicator establishes that distillation preserves both performance and uncertainty estimation \citep{guo2017calibration}.

\textbf{(2) Accurate but overconfident teacher.}  
The student improves calibration while preserving accuracy in ResNet152V2, indicating that mixup acts as a regularisation mechanism.
For ResNet152V2, the teacher achieves competitive accuracy but exhibits clear overconfidence. The student is able to maintain improvements in both calibration and accuracy. This suggests that mixup not only acts as a regularisation mechanism but also corrects the teacher's overconfident predictions via vicinal distributions.

\textbf{(3) Weak teacher.}  
For ConvNeXt-Tiny, the teacher has lower accuracy but relatively stable calibration. 
The student is unable to significantly improve accuracy beyond the teacher's, indicating a limitation in the transferable information. However, calibration is preserved, highlighting a Mixup stabilised effect.


\begin{table}
\centering
\small
\caption{
  Global and peak non-linearity residual alongside teacher ECE on
  CIFAR-100. Higher non-linearity correlates with a worse teacher
  calibration, with the exception of ViT-Base/16, which exhibits
  high non-linearity despite moderate calibration.
}
\begin{tabular}{lccc}
\toprule
Teacher & Global $\mathcal{L}_\text{NL}$ & Peak $\mathcal{L}_\text{NL}$ & ECE \\
\midrule
ConvNeXt-Large & 2.02 & 3.65 & 0.005 \\
ViT-Base/16    & 2.26 & 4.05 & 0.017 \\
ResNet152V2    & 2.61 & 4.68 & 0.062 \\
ConvNeXt-Tiny  & 1.72 & 3.10 & 0.029 \\
\bottomrule
\end{tabular}
\label{tab:lnl_summary}
\end{table}
\subsection{Teacher Non-Linearity under Vicinal Inputs}
\label{subsec:nonlinearity}
 
A teacher trained with standard empirical risk minimisation learns highly non-linear decision boundaries. We measure the deviation from this ideal linear model via the \emph{non-linearity residual} equation~\ref{eq:lnl}. 
$\mathcal{L}_\text{NL}$ increases as the teacher's response to $\tilde{x}$ deviates from the convex combination of its individual predictions.
 
\begin{figure}
    \centering
    \def\svgwidth{0.65\columnwidth}
    \begingroup
        \fontsize{5}{5}\selectfont
        \import{images/}{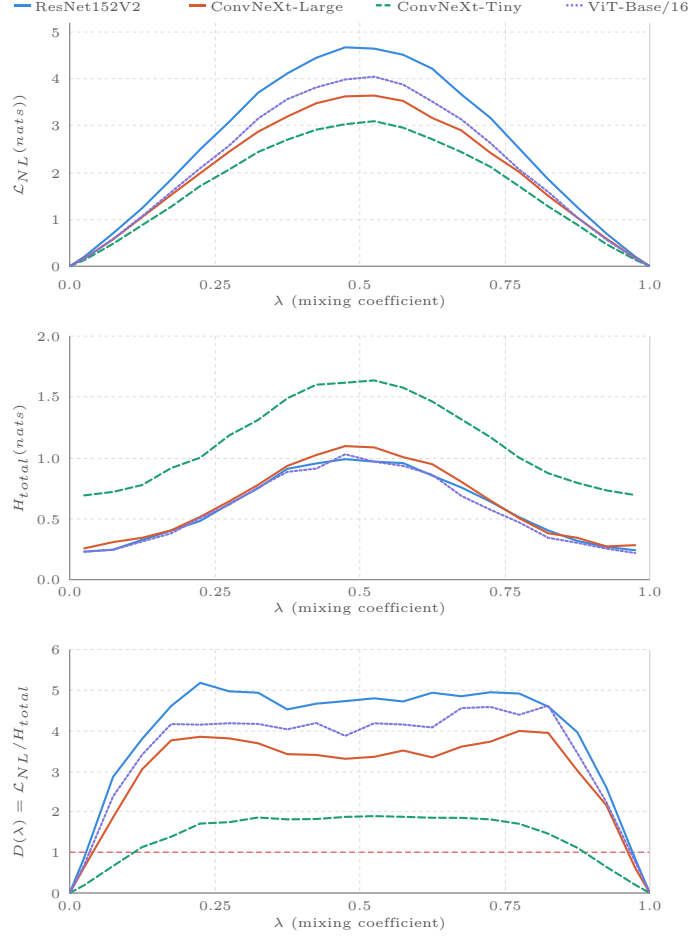}
    \endgroup
      \caption{
        Teacher behaviour on interpolated inputs $\tilde{x}$ for CIFAR-100 across mixing coefficient $\lambda \in (0,1)$.
    \textit{Top:} Non-linearity residual $\mathcal{L}_\text{NL}(\lambda)$, measuring deviation from linear behaviour between training examples; all teachers peak at $\lambda \approx 0.5$.
    \textit{Middle:} Total prediction entropy $H[p_t(\tilde{x})]$; ConvNeXt-Tiny is consistently more uncertain than ResNet152V2 and ConvNeXt-Large across $\lambda$ range.
    \textit{Bottom:} Dominance ratio $\mathcal{D}(\lambda) = \mathcal{L}_\text{NL} / H[p_t(\tilde{x})]$, above $\mathcal{D} = 1$ indicates mismatch dominates the supervisory signal. ResNet152V2 and ConvNeXt-Large exceed this threshold throughout the central interpolation region, while ConvNeXt-Tiny remains near with higher uncertainty.}
    \label{fig:entropy}
\end{figure}

Figure~\ref{fig:entropy} reports $\mathcal{L}_\text{NL}(\lambda)$ computed on the CIFAR-100 training set for all four teacher architectures. All curves are symmetric around $\lambda = 0.5$, where the mismatch is maximal, and vanish at the boundaries $\lambda \to 0$ and $\lambda \to 1$, where the interpolated input collapses to a real sample.
The ranking \mbox{ResNet152V2 $>$ ViT-B/16 $>$ ConvNeXt-L $>$ ConvNeXt-T} is consistent across the full $\lambda$ range, with global means of $2.61$, $2.02$, and $1.72$ respectively (Table~\ref{tab:lnl_summary}).

Notably, this ranking does not strictly follow architectural capacity but instead correlates with teacher calibration: ResNet152V2 is low calibrated and exhibits the largest non-linearity, whereas ConvNeXt-Tiny, despite being the weakest teacher, produces the smoothest response on interpolated inputs. A moderate calibration, such as ViT-Base/16, can also reflect a large non-linearity. 

\subsection{Signal Quality Decomposition}
\label{subsec:signal}
The teacher's non-linearity directly affects the quality of the supervisory signal received by the student. To quantify this, we use the dominance ratio $\mathcal{D}(\lambda)$ defined in equation~\ref{eq:dominance}, which measures the fraction of the teacher's total prediction entropy on $\tilde{x}$ attributable to distributional mismatch rather than to the natural ambiguity of the mixed label.
 
Figure~\ref{fig:entropy} shows that all three teachers exhibit $\mathcal{D}(\lambda) > 1$ near $\lambda = 0.5$ 
. This confirms that the supervisory signal in the high-interpolation regime
is dominated by distributional mismatch rather than by informative
inter-class structure.
Crucially, ConvNeXt-Tiny, the least confident teacher overall
($H_\text{global} = 1.13$ nats), is least affected by this phenomenon,
suggesting that teacher overconfidence amplifies the mismatch effect.
 
\subsection{What the Student Learns Beyond Imitation}
\label{subsec:student}
 
Having characterised the teacher signal, we now ask whether the student
merely imitates the teacher or develops properties the teacher does not
possess. For each teacher--student pair, we compute $\mathcal{L}_\text{NL}$ for the student using the same protocol as equation~\ref{eq:lnl}, and using the \emph{linearity gain} in equation \ref{eq:lingain_info}. Positive gains in $\Delta_\text{lin}$ indicate that the gain student linearity with respect to its teacher on $\tilde{x}$, a property the teacher was never trained to have.
 
\begin{figure} [h]
    \centering
    \def\svgwidth{0.65\columnwidth}
    \begingroup
        \fontsize{5}{5}\selectfont
        \import{images/}{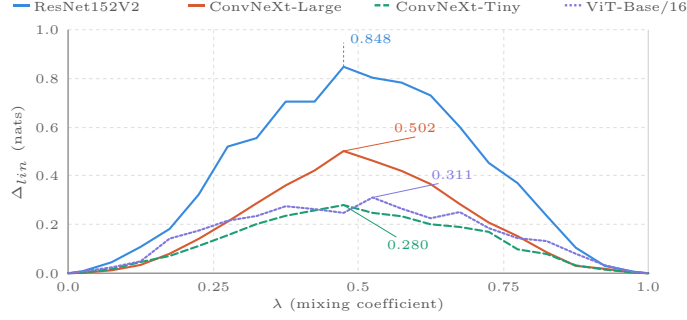}
    \endgroup
  \caption{
    Linearity gain $\Delta_\text{lin}(\lambda)$ for the three
    teacher--student pairs.
    In all cases the student is more linear than its teacher across the
    full $\lambda$ range.
    The gain is proportional to the teacher's non-linearity and peaks at
    $\lambda \approx 0.5$.
  }
  \label{fig:lingain}
\end{figure}

Figure~\ref{fig:lingain} shows that $\Delta_\text{lin}(\lambda) > 0$ across all $\lambda$ and for all teacher--student pairs.
The gain peaks near $\lambda = 0.5$ with values of $+0.848$ (ResNet152V2), $+0.502$ (ConvNeXt-L), $0.311$ (ViT-B/16), and $+0.28$ (ConvNeXt-Tiny),
following the same ordering as $\mathcal{L}_\text{NL}^{(t)}$.
This proportionality is not coincidental: the more non-linear the teacher on vicinal inputs, the stronger the linearising effect on the student.

To assess whether this linearity gain reflects genuine independent structure or mere divergence from the teacher, we examine the alignment $D_\text{KL}(p_t(\tilde{x}) \| p_s(\tilde{x}))$ evaluated separately on original (real) samples $x$ and on interpolated samples $\tilde{x}$ (Table~\ref{tab:alignment}). Low $\text{KL}_\text{real}$ confirms that the student faithfully learned from the teacher on the clean domain where the teacher is reliable. The substantially higher $\text{KL}_\text{mixed}$ reveals that on vicinal inputs, precisely where the teacher is most non-linear and least reliable,  the student diverges and develops its own response. Rather than inheriting the teacher's confusion in the interpolated region, the student generalises beyond imitation.
 
\begin{table}[h]
\centering
\small
\caption{Peak linearity gain and teacher--student distributional alignment on original vs.\ interpolated inputs. $\text{KL}\text{real}$ evaluated on clean samples $x$ measures how closely the student learned from the teacher in the domain where the teacher is reliable. $\text{KL}\text{mixed}$ evaluated on vicinal samples $\tilde{x}$ measures their agreement in the interpolated region. The gap $\text{KL}\text{mixed} - \text{KL}\text{real}$ reflects the degree to which the student developed an independent response to mixup rather than imitating the teacher. ViT-Base/16 shows the largest such gap despite the lowest $\text{KL}_\text{real}$, indicating the strongest independence from the teacher on vicinal inputs.}
\begin{tabular}{lcccc}
\toprule
Teacher & Peak $\Delta_\text{lin}$ & $\text{KL}_\text{real}$ & $\text{KL}_\text{mixed}$ & Indepen. Response \\
\midrule
ResNet152V2    & $+0.825$ & $0.071$ & $0.469$ & $0.398$ \\
ConvNeXt-Large & $+0.482$ & $0.039$ & $0.320$ & $0.281$ \\
ViT-Base/16    & $+0.279$ & $0.040$ & $0.821$ & $0.781$ \\
ConvNeXt-Tiny  & $+0.264$ & $0.055$ & $0.216$ & $0.161$ \\
\bottomrule
\end{tabular}
\label{tab:alignment}
\end{table}
 
\subsection{Robustness Transfer under Distribution Shift}
\label{subsec:robustness}
 
The linearity induced by mixup is not confined to the interpolation manifold: it generalises to corrupted distributions.
We show the results of evaluating teacher--student pairs on CIFAR-100-C, reporting accuracy gain over the mixup-only baseline student $\Delta_\text{acc} = \text{acc}_\text{student} - \text{acc}_\text{baseline}$ across six corruption types and five severity levels.
\begin{table}
\centering
\small
\caption{Accuracy (\%) under common corruptions at different severity 
levels on CIFAR-100-C (Bold indicates the best student performance 
per corruption and severity).}
\begin{tabular}{llccccc}
\toprule
\multicolumn{7}{c}{\textbf{Student MobileNetV2}} \\
\midrule
Teacher & Corruption & S1 & S2 & S3 & S4 & S5 \\
\midrule
\multirow{6}{*}{\textbf{Baseline}}
 & gaussian\_noise & 33.78 & 18.52 & 10.71 & 7.90  & 6.06  \\
 & motion\_blur    & 74.02 & 66.17 & 56.52 & 56.68 & 47.07 \\
 & fog             & 80.21 & 77.71 & 73.61 & 66.66 & 46.90 \\
 & jpeg            & 63.49 & 55.33 & 51.53 & 47.90 & 42.07 \\
 & contrast        & 79.72 & 73.81 & 67.22 & 56.13 & 26.87 \\
 & pixelate        & 76.38 & 67.55 & 61.07 & 40.02 & 19.45 \\
\midrule
\multirow{6}{*}{\textbf{ConvNeXt-L}}
 & gaussian\_noise & 47.72 & 26.83 & 15.38 & 11.97 & 9.27  \\
 & motion\_blur    & \textbf{81.82} & \textbf{76.19} & 68.27 & \textbf{68.72} & \textbf{60.41} \\
 & fog             & \textbf{86.13} & \textbf{84.99} & \textbf{83.47} & \textbf{80.15} & \textbf{64.85} \\
 & jpeg            & 69.48 & 59.20 & 55.81 & 51.31 & 45.57 \\
 & contrast        & \textbf{86.04} & \textbf{84.39} & \textbf{82.88} & \textbf{79.94} & \textbf{65.88} \\
 & pixelate        & \textbf{81.82} & \textbf{70.66} & \textbf{63.64} & 37.78 & 18.05 \\
\midrule
\multirow{6}{*}{\textbf{ViT-B/16}}
 & gaussian\_noise & 46.39 & 26.03 & 14.79 & 11.52 & 9.11  \\
 & motion\_blur    & 81.62 & 76.38 & \textbf{69.24} & 69.53 & 61.87 \\
 & fog             & 85.66 & 84.22 & 81.82 & 77.05 & 59.92 \\
 & jpeg            & \textbf{71.51} & \textbf{62.45} & \textbf{59.07} & \textbf{55.85} & \textbf{49.91} \\
 & contrast        & 85.63 & 83.73 & 80.97 & 75.74 & 50.00 \\
 & pixelate        & 81.42 & 70.66 & 63.64 & 37.78 & 18.05 \\
\midrule
\multirow{6}{*}{\textbf{ResNet152V2}}
 & gaussian\_noise & \textbf{49.18} & \textbf{29.12} & \textbf{18.35} & \textbf{14.64} & \textbf{12.05} \\
 & motion\_blur    & 78.83 & 72.13 & 62.94 & 63.77 & 54.18 \\
 & fog             & 83.61 & 81.79 & 78.82 & 73.92 & 56.03 \\
 & jpeg            & 70.21 & 61.07 & 57.22 & 53.76 & 48.70 \\
 & contrast        & 83.61 & 80.70 & 76.90 & 69.31 & 38.24 \\
 & pixelate        & 79.29 & 68.86 & 60.17 & 35.74 & 18.16 \\
\midrule
\multirow{6}{*}{\textbf{ConvNeXt-T}}
 & gaussian\_noise & 41.28 & 24.53 & 15.60 & 12.48 & 10.64 \\
 & motion\_blur    & 76.03 & 69.83 & 60.24 & 60.85 & 52.02 \\
 & fog             & 80.59 & 78.85 & 76.07 & 71.65 & 53.18 \\
 & jpeg            & 63.68 & 55.34 & 51.87 & 48.63 & 44.15 \\
 & contrast        & 80.49 & 77.83 & 75.29 & 70.33 & 51.25 \\
 & pixelate        & 76.64 & 64.44 & 55.57 & 30.06 & 14.07 \\
\bottomrule
\end{tabular}
\label{tab:corruptions}
\end{table}

\begin{figure} [t]
    \centering
    \def\svgwidth{0.7\columnwidth}
    \begingroup
        \fontsize{5}{5}\selectfont
        \import{images/}{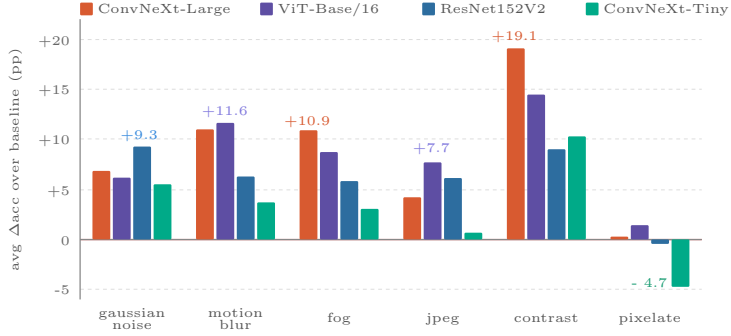}
    \endgroup
  \caption{Average accuracy gain over the baseline student on CIFAR-100-C, grouped by corruption type. The ranking across teachers mirrors the ordering of linearity gains. Fog and contrast benefit most; pixelate is anomalous at high severity.}
  \label{fig:corruption_gain}
\end{figure}

Table~\ref{tab:corruptions} and Figure~\ref{fig:corruption_gain} show
that the average gain follows the same ranking as the linearity gain:
ConvNeXt-L $>$ ViT-B/16 $>$ ResNet152V2 $>$
ConvNeXt-Tiny.
Smooth corruptions such as fog and contrast benefit most, which is consistent with mixup producing smooth interpolations between training
examples - the inductive bias introduced during distillation directly covers this type of perturbation.
 
Gaussian noise, in contrast, introduces high-frequency perturbations that are structurally different from convex combinations of images.
The gain is smaller and degrades rapidly with severity, confirming that mixup-induced linearity does not provide universal robustness but is
selective to perturbations that are geometrically similar to interpolations.
 
Pixelate represents a systematic exception that affects all four distilled students, with degradation onset varying by teacher strength. Concretely, the ConvNeXt-Tiny student falls below the baseline from severity~2 onwards, ResNet152V2 from severity~3, ViT-B/16 and ConvNeXt-Large from severity~4. This pattern is consistent with the inductive bias introduced by mixup. Convex combinations of images produce smooth, low-frequency blends that reinforce sensitivity to global spatial structure. Pixelate at high severity destroys precisely this structure by replacing local regions with uniform colour blocks, creating a distribution that is orthogonal to the interpolation manifold. 

As a result, all distilled students, trained to exploit smooth spatial transitions, are more exposed to this corruption than the baseline student trained without mixup supervision. The ConvNeXt-Large student is most resilient at low severity because its teacher transfers richer representations, but this advantage erodes at high severity where the corruption overwhelms any structural prior.
 
Taken together, these results establish that mixup-based distillation transfers more than soft-label supervision.
The student acquires a structural property, greater linearity in the vicinal region, that its teacher does not possess, and this property manifests as improved robustness under smooth distribution shifts.
 
\subsection{Discussion}
\label{subsec:discussion}
 
The results reported collectively address the three claims of the hypothesis stated in Section~\ref{sec:hypothesis}.
We discuss each in turn before drawing the overall conclusions of
this section.
 
\paragraph{Claim 1: the teacher's supervisory signal is dominated by
distributional mismatch.}
This claim is confirmed by the dominance ratio $\mathcal{D}(\lambda) > 1$ near $\lambda = 0.5$ for all three
teacher architectures (Section~\ref{subsec:signal}).
In the high-interpolation regime, the mismatch term $\mathcal{L}_\text{NL}(\lambda)$ exceeds the total prediction entropy $H[p_t(\tilde{x})]$, meaning that the teacher's confusion about the vicinal input outweighs any informative inter-class structure it could otherwise provide.
From the information-theoretic perspective of Section~\ref{subsubsec:mi}, the lower bound on $I(p_t; p_s)$ grows with $\mathcal{L}_\text{NL}$ but this growth is not a guarantee of more \textit{useful knowledge}.
 
\paragraph{Claim 2: the student acquires greater linearity
independently of the teacher.}
This claim is confirmed by the linearity gain
$\Delta_\text{lin}(\lambda) > 0$ across all $\lambda$ and all
teacher--student pairs (Section~\ref{subsec:student}).
For these the gain even has peaks near $\lambda = 0.5$, thus, the more non-linear the teacher, the
stronger the regularisation mixup exerts on the student. This reflects the fact that the vicinal training objective penalises non-linear behaviour precisely where the teacher's mismatch is largest. The fourfold degradation in teacher--student alignment on mixed inputs compared to real inputs confirms that this linearity is not transferred from the teacher but developed independently by the student. This structural property lies beyond what the mutual information lower bound of equation~\ref{eq:mi_bound} can account for under standard KD.

\paragraph{Claim 3: the induced linearity generalises beyond the
interpolation manifold.}
This claim is confirmed by the robustness results on CIFAR-100-C (Section~\ref{subsec:robustness}).
The average accuracy gain over the baseline student follows the same ranking as the linearity gain.
Smooth corruptions benefit most, consistent with mixup producing smooth interpolations between training examples
whose inductive bias directly covers this type of perturbation. High-frequency perturbations structurally different from convex combinations of images yield smaller and severity-dependent gains, confirming that the
\textit{generalisation is selective rather than universal}.
 
\paragraph{Limits and trade-offs.}
The one systematic limit of mixup is the degradation of fine spatial details at high severity, where all distilled students fall below the baseline, with the onset of degradation varying by teacher strength. This reveals an inherent trade-off in vicinal training. This trade-off is not specific to any teacher architecture; it is a
consequence of the vicinal training objective itself and represents an open direction for future work.

\section{Conclusion}

In this work, we studied the interaction between Knowledge Distillation and Mixup in a setting where mixup is applied only during student training and the teacher is queried on interpolated inputs drawn from a vicinal distribution. This setup introduces a controlled distribution mismatch that allows for analysing how input-level smoothing affects the quality of the teacher signal and its transfer to the student.

Our experiments on CIFAR-10, CIFAR-100, and ImageNet show that KD with mixup consistently improves over the baseline student in accuracy and, in most cases, also yields substantially better calibration. Calibration properties propagate through the distillation process in a direction that accuracy alone cannot predict. A poorly calibrated teacher does not merely limit student performance: it actively shapes the student's uncertainty profile, potentially amplifying overconfidence or inducing underconfidence even when accuracy transfer is successful.

The effect of distillation temperature is not independent of mixup: together they operate as interacting smoothing mechanisms whose combined effect on the accuracy--calibration trade-off is more pronounced than either alone. Moderate temperatures preserve the informative structure of the teacher signal under vicinal inputs, while higher values over-smooth the supervisory signal at the cost of discriminative performance.

These findings collectively challenge the standard view of KD as a fidelity-maximisation process. The student's gains arise not from closer imitation of the teacher but from acquiring an inductive bias, linearity in the vicinal region, that the teacher itself lacks and that the mutual information framework of standard KD cannot account for. This suggests that vicinal training opens a richer channel of knowledge transfer than soft-label supervision alone, one that warrants explicit characterisation in future distillation frameworks.

Overall, our study provides empirical evidence that the benefits of mixup in distillation depend on a balance between teacher quality, student capacity, and smoothing strength. These findings offer practical guidance for designing distillation pipelines and suggest that accuracy and calibration should be treated as distinct objectives when selecting temperature and teacher architecture under vicinal training.

\section*{Acknowledgment}
The authors acknowledge the support of the European Union and Estonian Research Council via project TEM-TA101 and the collaboration project LLTAT21278 with Bolt Technologies.

\bibliography{refs}
\newpage
\appendix
\section{Teacher Capacity and Knowledge Transfer}
\label{app:capacity_smootnes}

A common assumption is that higher-capacity teachers provide better supervision \citep{gou2021knowledge}. While larger models indeed learn more discriminative representations, this does not necessarily translate into knowledge transfer \citep{wang2021knowledge,huang2022knowledge, zhang2016understanding}. Figure~\ref{fig:vscore} illustrates the feature representations extracted from the penultimate layer for similar CIFAR-100 classes (baby, boy, girl, man, woman) across different architectures. Under ERM (figure~\ref{fig:vscore}-a), clustering quality improves significantly with model capacity, as reflected by higher V-scores and clearer class separation. This indicates that stronger models learn highly discriminative features.
However, this increased separability leads to sharper decision boundaries and more confident predictions; see the green decision boundary in figure \ref{fig:emp_visi} for ERM. In this regime, the teacher's output distributions approach one-hot vectors, reducing the information content of soft targets \citep{guo2017calibration, hinton2015distilling}. As a result, inter-class relationships, crucial for effective distillation, become less expressive.

\subsection{Dark Knowledge Smoothness}
\begin{figure}
    \centering
    \def\svgwidth{0.72\columnwidth}
    \begingroup
        \fontsize{4}{4}\selectfont
        \import{images/}{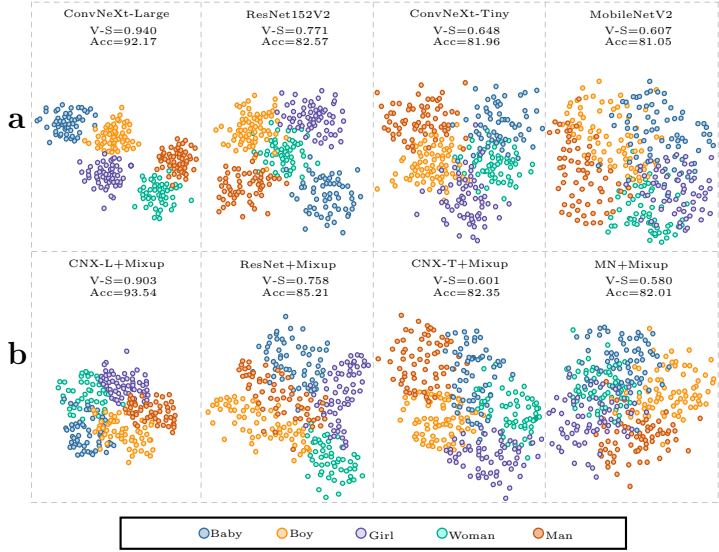}
    \endgroup
    \caption{Penultimate-layer feature representations on CIFAR-100 for five related classes, visualised with t-SNE. Columns correspond to DNN architectures, and rows to training regimes: a) ERM and b) VRM mixup. V-score indicates clustering quality and Acc classification accuracy. While ERM yields compact, well-separated clusters, mixup introduces smoother class transitions.}
    \label{fig:vscore}
\end{figure}

Figure~\ref{fig:vscore}-b  shows  mixup training as VRM. Here, decision boundaries become smoother, transitioning from sharp separations to continuous regions.  In figure \ref{fig:emp_visi} (VRM with mixup), the boundary decision is a green region, not a line. Although this reduces clustering compactness (lower V-score), it improves accuracy reflecting better generalisation. Mixup implicitly introduces soft labels and encourages the model to capture intermediate class relationships \citep{choi2023understanding}. Similarly, KD introduces softness by temperature-scaling the logits \citep{hinton2015distilling}. However, an optimal temperature becomes highly dependent on the teacher’s confidence and architecture, making it a sensitive hyperparameter.
\begin{figure}
    \centering
    \def\svgwidth{0.72\columnwidth}
    \begingroup
        \fontsize{4}{4}\selectfont
        \import{images/}{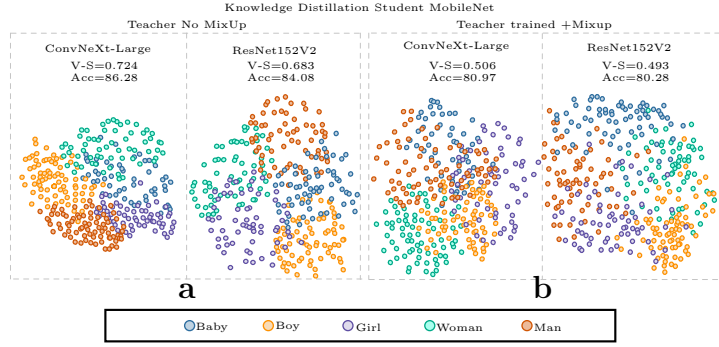}
    \endgroup
    \caption{Penultimate-layer feature representations on CIFAR-100 for five related classes, visualised with t-SNE. Columns correspond to two different student-teacher configurations: a) Students trained with teachers without previous mixup, and b) Students trained with teachers with a previous mixup. For each case, we use two teachers, Convnext-Large and Resnet152V2. V-score indicates clustering quality and Acc classification accuracy. Figure b shows a clear effect of excessive smoothing in distillation, degrading both clustering structure and performance.}
    \label{fig:vscore2}
\end{figure}

The interaction between these effects becomes evident in the figure~\ref{fig:vscore2}, where a student is trained via KD under different teacher regimes. When distilling from teachers trained without mixup figure~\ref{fig:vscore2}-a, the student benefits from a balanced level of softness: smooth supervision is introduced only through temperature scaling and mixup applied at the student. This results in well-structured representations and strong performance.

However, when the teacher is also trained with mixup figure~\ref{fig:vscore2}-b, a compounding effect emerges. The student is simultaneously exposed to three sources of softening: (i) mixup in the teacher training, (ii) mixup during student training, and (iii) temperature scaling in distillation. This excessive smoothing degrades both clustering structure and accuracy, leading to representations that are less discriminative than even standard training.

These observations highlight a critical trade-off between teacher capacity, representation sharpness, and supervision quality. In figure \ref{fig:emp_visi}, we can see that KD and mixup-based distillation seem to have a balance in the decision boundary region. While high-capacity teachers improve discrimination, they may reduce the transferability of knowledge. Conversely, excessive smoothing-arising from multiple sources-can harm representation learning. This motivates the need for controlled mechanisms that balance discrimination and smoothness during distillation.
\section{Effect of Distillation Temperature}
\label{sec:temp_effects}
\begin{figure}
    \centering
    \def\svgwidth{1\columnwidth}
    \begingroup
        \fontsize{5}{5}\selectfont
        \import{images/}{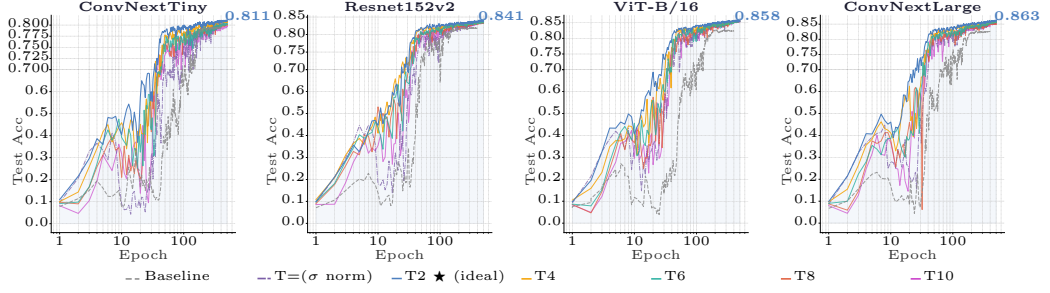}
    \endgroup
    \caption{Test accuracy showed by teacher architecture trained with different temperatures
    over 500 epochs (log scale). $T{=}2$ consistently achieves the highest
    accuracy throughout training, reaching the upper bound earliest, and
    maintaining a slight but persistent advantage over the remaining
    configurations. $T_\text{std}$ exhibits erratic behaviour during the early
    epochs before recovering and stabilising in the later phases, ultimately
    approaching but not surpassing $T{=}2$. The vertical axis applies a
    piecewise linear scale above $0.7$ ($5\times$ expansion) to enhance
    visibility of late-training differences.}
    \label{fig:acc_temperature}
\end{figure}

We investigate the effect of the distillation temperature $T$ on the student's learning dynamics. Several temperature configurations are evaluated: baseline KD with $T=2$, mixup KD with fixed temperatures $T \in \{2, 4, 6, 8, 10\}$ and an adaptive variant proposed by \citet{choi2023understanding} where $T$ is set to the standard deviation of the teacher's logits ($T_{\sigma}$), effectively normalising the soft target distribution at each training step. Figure~\ref{fig:acc_temperature} shows test accuracy over 500 epochs for each architecture on a logarithmic time axis, which better reveals the dynamics of both the early and late training phases. 

The trajectories differ substantially. $T{=}2$ reaches high accuracy earliest and maintains a consistent, if small, advantage throughout training, supporting its selection as the default configuration in our experiments. For temperatures above $T{=}4$, the final accuracy values converge to a narrow range ($\sim$0.5--0.8\%), suggesting that the choice of temperature has limited impact on the ultimate performance ceiling. Nevertheless, $T{=}4\rightarrow10$ exhibit a consistent trend: accuracy decreases monotonically with increasing temperature, an effect explained by the progressive smoothing of the supervision signal, which dilutes class discriminability as temperature grows. 

The adaptive variant $T_{\sigma}$ presents a more nuanced picture. During the first 50 epochs, it falls visibly below all fixed-temperature configurations,
exhibiting erratic behaviour that we attribute to the instability of logit variance in the early stages of training, when the student has not yet learned
meaningful representations. Despite this slow start, $T_{\sigma}$ recovers and stabilises in later phases, approaching the accuracy of $T{=}2$ by the end
of training. This suggests that adaptive normalisation is a viable strategy given a sufficient training budget, but offers no clear advantage over a fixed temperature in the regime studied here.

\end{document}